\pdfoutput=1

\documentclass[11pt]{article}

\usepackage{acl}

\usepackage{times}
\usepackage{latexsym}

\usepackage[T1]{fontenc}

\usepackage[utf8]{inputenc}

\usepackage{microtype}

\usepackage{inconsolata}

\usepackage{graphicx}
\usepackage{multirow}
\usepackage{multicol}
\usepackage{amssymb}
\usepackage{amsmath}
\usepackage{tabulary}
\usepackage{booktabs}
\usepackage{enumitem}
\usepackage{subcaption}

\usepackage{makecell}

\usepackage{amsthm}
\usepackage{tcolorbox}

\usepackage{xspace}
\newcommand{\dataset}{\textsc{CPopQA}\xspace} 

\title{\dataset: Ranking Cultural Concept Popularity by LLMs }

\author{Ming Jiang\and  Mansi Joshi  \\
        Department of Human-Centered Computing \\ 
        Indiana University, Indianapolis, IN, USA\\
        \texttt{mj200@iu.edu, joshma@iu.edu}}

\begin{document}
\maketitle

\begin{abstract}
    Prior work has demonstrated large language models' (LLMs) potential to discern statistical tendencies within their pre-training corpora. Despite that, many examinations of LLMs' knowledge capacity focus on knowledge explicitly appearing in the training data or implicitly inferable from similar contexts. How well an LLM captures the corpus-level statistical trends of concepts for reasoning, especially long-tail ones, is still underexplored. In this study, we introduce a novel few-shot question-answering task (CPopQA) that examines LLMs' statistical ranking abilities for long-tail cultural concepts (e.g., holidays), with a specific focus on these concepts' popularity in the United States and the United Kingdom, respectively. We curate a dataset containing 459 holidays across 58 countries, generating a total of 6,000 QA testing pairs. Experiments on four strong LLMs show that large models are capable of ranking long-tail cultural concepts regarding their statistical tendency. Notably, GPT-3.5 displayed superior performance and exhibited its potential to identify geo-cultural proximity across continents.

\end{abstract}
\section{Introduction}
\label{sec:introduction}
Recent large language models (LLMs) have shown their potential to capture multiple facets of the world, offering advantages to a variety of downstream applications such as constructing knowledge bases with reduced reliance on human intervention \cite{knowledgebase_1,knowledgebase_2}. However, the boundary of LLMs' capabilities is still an open question, causing uncertainty in practical LLM deployment.

To advance LLM understanding, researchers have been actively evaluating these models on various knowledge-intensive tasks, encompassing diverse aspects such as factual knowledge~\cite{factual1}, commonsense~\cite{commonsense1}, basic science~\cite{science}, and simple math ~\cite{math}. Given the generative nature of LLMs, these tasks can be easily conducted in a question-answering (QA) format. Early studies in this thread emphasize LLMs' memory capacity and discover that LLMs possess a remarkable ability to embed linguistic phenomenon~\cite{llmsyn,linguisticprob}, language's statistical tendencies \cite{lmstatis1,llmstatis2,llmstatis3}, and geo-cultural commonsense \cite{geocommonsense}. Recently, advances in prompting like chain-of-thought \cite{chainofthought,cot3} and self-reflection \cite{selfreflexion} have enabled LLMs to elicit complex multi-hop reasoning from in-context examples. Beyond common knowledge, the latest studies further explore LLMs' ability to process long-tail knowledge \cite{longtail}, particularly in underserved NLP domains, e.g., humanities \cite{humanities} and cultural analytics \cite{cultural,cult2}. 

Despite insightful findings, existing examinations largely focus on the capability of LLMs to grasp knowledge explicitly appearing in the training data or implicitly inferable from similar contexts. There is scant research into the capacity of LLMs to \textit{capture the broad statistical patterns of concepts within extensive datasets for in-depth comparisons, especially long-tail concepts spanning significantly diverse sociocultural contexts}. This alternative perspective focuses on models' potential to embed macro-level phenomena derived from widely scattered knowledge points in the training corpus, which can broaden LLMs' benefits as exploratory tools in support of corpus-centered computational analysis \cite{statistical_text}, such as helping digital humanists and social scientists to gain new insights into historical, cultural, and social problems \cite{dan_css,culture_ted}.

In this study, we attempt to explore the statistical ranking ability of LLMs, with a specific focus on a research question: \textit{Can large language models compare cultural concepts, especially long-tail ones, regarding their worldwide popularity?} To examine this question, we design a ranking-based statistical QA task that compares cultural concept popularity across countries (called CPopQA). To support this study, we curate a benchmark dataset of 6,000 QA testing pairs, covering 459 holidays across 58 countries. Note that, our dataset construction process is flexible and scalable, allowing for the easy generation of diverse testing instances. Experiments on four popular LLMs show that the large models are capable of ranking holidays regarding their statistical tendency. In particular, GPT-3.5 outperformed other models and showed a potential to identify geo-cultural proximity across continents.

\section{\dataset}
\label{sec:dataset}
In this section, we introduce our \dataset by describing the task formulation, the process of dataset construction, and a prompt-based LLM approach.

\paragraph{Task} Considering the geo-association between holidays and countries, we define a ranking-based statistical QA task regarding worldwide holiday popularity as below:

\vspace{-0.5em}
\begin{itemize}[leftmargin=10pt] \itemsep-0.2em
    \item Given a set of holidays $\mathcal{H} = \{h_1, h_2, \dots, h_n\}$ from a query country $c_q$, the goal is to sort $\mathcal{H}$ in a descending order based on their popularity in a target country $c_t$.
\end{itemize}
\vspace{-0.5em}

\paragraph{Holiday List Curation} To create the QA dataset, we started by curating global holidays from Wikipedia's list of public holidays by country\footnote{\url{https://en.wikipedia.org/wiki/Category:Lists\_of\_public\_holidays\_by\_country}}, considering the following factors:

\vspace{-0.5em}
\begin{enumerate}[leftmargin=15pt] \itemsep-0.15em
 \item \textbf{Diversity and inclusivity across geo-cultures:} 
 we considered the holidays from both under-represented and well-represented countries. Specifically, we referred to the population statistics and the number of holidays on each country's wiki page, and collected holidays from the top five and bottom five countries regarding population statistics in each continent.
 
 \item \textbf{Valid wiki page:} We required the selected holidays to have valid wiki pages to guarantee the authenticity of collected items. Meanwhile, we extracted the first paragraph from each holiday's wiki page as its description. This enables the future development of methods using text descriptions of these holidays.
 
 \item \textbf{Date variation} We curated the holiday list by adding their countries and dates because many common holidays across countries are celebrated on different dates. For example, \textit{Labor Day} is celebrated on \textit{May 1st in China}, whereas it falls on \textit{September 5th in the United States}.
\end{enumerate}
\vspace{-0.7em}
Due to the editions of different crowd workers and the unique characteristics of holidays, we conducted a series of data cleaning (Appendix \ref{sec:datacleaning}). 

\paragraph{Holiday Popularity Collection} We estimate the overall engagement per holiday among people based on the average frequency of each holiday's name over $\sim$220 years (1800--2019), counted by Google Books Ngram Viewer\footnote{\url{https://books.google.com/ngrams/}} (GBNV). This tool has been widely used to analyze user-selected word or phrase frequency. The corpus of GBNV consists of digitized books, which, to our best knowledge, is the largest public collections of books across a wide spectrum of domains and time periods \footnote{Over 6\% of all books ever published in the 2012 version \cite{ggngram2}, and we use the updated 2020 version.}, making it affordable with representative n-gram statistics. Moreover, GBNV provides several sub-corpora categorized by the books’ publication country, making it an ideal resource to collect worldwide holiday popularity within a specific country. Given that English is the most accessible language denoting various worldwide holidays, especially for holidays from non-English speaking countries, we estimate holiday popularity based on two English sub-corpora (i.e., American and British English corpus), respectively. 

To validate our strategy for deriving holiday popularity, we further conducted a human evaluation with 6 annotators (details in Appendix \ref{sec:humaneval}). Our results show that the GBNV statistics achieved $\sim$60\% consistency with human judgments on average. In total, we collected information on 459 holidays in 58 countries on 5 continents. Each holiday is annotated with its country, date, description, and frequency in American/British corpora. The details of the data statistics are in Appendix \ref{sec:datastat}.

\paragraph{QA Pair Construction} To investigate the influence of ranking complexity on model performance, we constructed questions to rank $n$ items for both tasks, where $n\in\{2,3,5\}$. For example in Task 1, we sampled a holiday set $\mathcal{H}=\{h_1, h_2, \dots, h_n\}$ in a query country $c_q$ from our complete holiday list, and sorted them by their popularity in a target country $c_t$ (e.g., US or UK) to get a ranked list $[h'_1, h'_2, \dots, h'_n]$. We followed prior work~\cite{geocommonsense} by using either the country names (e.g., holidays in Nigeria) or their corresponding modifiers (e.g., Nigerian holidays) to denote the query countries in the questions.  We then selected the optimal question template with the ranked holiday list as the answer for further analysis:

\begin{table}[]
    \centering
    \resizebox{0.75\columnwidth}{!}{%
    \begin{tabular}{ccc}
    \toprule
     \multicolumn{1}{l}{Setting}  & \multicolumn{1}{l}{\#Countries} & \multicolumn{1}{l}{\#Holidays} \\
    \midrule
    2-item ranking & 58 & 255 \\
    3-item ranking & 57 & 265 \\
    5-item ranking & 55 & 271 \tabularnewline 
    \bottomrule   
    \end{tabular}
    }
    \vspace{-0.5em}
    \caption{Holiday Diversity in Testing Data.}
    \vspace{-1.5em}
    \label{tab:dataset diversity}
\end{table}

\begin{tcolorbox}[boxrule=0pt, frame empty]
   Question~=~``Can you provide a descending order for the following $c_q$ holidays by their popularity in the $c_t$: $h_1, h_2, \dots, h_n$'' \\
   Answer~=~``$1. h'_1$,~~$2. h'_2$, ...,~~$n. h'_n$''
\end{tcolorbox}

For each $n\in\{2,3,5\}$ and each of the two tasks, we created 200 pairs for ranking $n$ items. To examine the variation of results, in each setting, we conducted five rounds of holiday set sampling for QA pair generation, and then we repeated all the experiments. Thus, our QA dataset consists of 6,000 QA pairs in total. Table~\ref{tab:dataset diversity} shows the holiday and country statistics in each ranking setting. Note that, we considered both US and UK as the target countries in this holiday ranking task. 

\paragraph{Prompting} Following \citet{longtail}, we used a simple prompt template: ``Q: [Question] \textbackslash n A:[Answer]'' and randomly selected 3 in-context examples \footnote{we tried different sizes of in-context examples (e.g., 2, 3, 5) and observed similar trends regarding model performance.} to form a prompt. Feeding the prompt to an LLM, we generated ranks by greedy decoding and we compared them with the ground truth.

\section{Experimental Setting}
\label{sec:qa}

\paragraph{LLMs and Baselines} We chose 4 popular LLMs for evaluation. The first LLM is \textbf{GPT-3.5} (i.e., text-davinci-003, 175B parameters). Through fine-tuning GPT-3 by reinforcement learning from human preferences \cite{gpt35}, GPT-3.5 shows a higher quality in handling complex instructions compared to prior GPT-based models. We next chose LLaMA, with 7B (\textbf{LLaMA-7B}) and 13B (\textbf{LLaMA-13B}) parameters, pre-trained on the English-dominated corpora covering diverse domains \cite{llama}. The final model is \textbf{BLOOM-7b1}, a multilingual LLM with $\sim$7B parameters \cite{bloom}. We selected this BLOOM variant because of its comparable model size with LLaMA-7B. We employed 3 baselines, including random guess and statistical simulation by Google Trends and Wikipedia article length, respectively (see details in Appendix~\ref{sec:baseline}).

\paragraph{Metrics} We used three evaluation metrics, including \textbf{Accuracy (Acc.)} measures the degree of the exact match; \textbf{Precision@1 (P@1)} calculates the precision of the first ranked item; \textbf{Average difference ($\textbf{Diff.}=\frac{1}{N}\sum_{j=1}^N \frac{1 - \rho_j}{2}$)} measures the overall ranking difference, where $\rho_j$ is the Spearman correlation coefficient between the model prediction and the ground-truth ranking on the $j$-th QA example.

\section{Results and Analysis}
\label{sec:results}

\begin{table}[t]
\centering
\resizebox{0.98\columnwidth}{!}{%
\begin{tabular}{llccc}
\toprule
Setting                          & Model  & \multicolumn{1}{l}{P@1 (\%)} & \multicolumn{1}{l}{Acc. (\%)} & \multicolumn{1}{l}{Diff.}    \\
\midrule
\multirow{7}{*}{2-item ranking} & random guess  & 50.00 $\pm$ 0.00 & 50.00 $\pm$ 0.00 & -     \\
                                 & google stat  & 57.00 $\pm$ 0.03 & 57.00 $\pm$ 0.03  & -     \\
                                 & wiki len  & \textbf{59.20} $\pm$ 0.04 & \textbf{59.20} $\pm$ 0.04  & -     \\
                                 & bloom-7b1 & 42.90 $\pm$ 0.04                        & 42.90 $\pm$ 0.04                      & -                            \\
                                 & llama-7b  & 48.20 $\pm$ 0.03                     & 48.20 $\pm$ 0.03                         & -                            \\
                                 & llama-13b  & 51.10 $\pm$ 0.05                        & 51.10 $\pm$ 0.05                           & -                            \\
                                 & gpt-3.5   & 54.80 $\pm$ 0.03                       & 54.80 $\pm$ 0.03                        & -                            \\
\cmidrule(lr){1-5}
\multirow{7}{*}{3-item ranking} & random guess  & 33.33 $\pm$ 0.00 & 16.67 
                                   $\pm$ 0.00  & 0.500 $\pm$ 0.00 \\
                                 & google stat  & 37.00 $\pm$ 0.03 & 21.40 $\pm$ 0.01  & 0.441 $\pm$ 0.01 \\
                                 & wiki len  & 53.5 $\pm$ 0.03 & 28.10 $\pm$ 0.03  & 0.378 $\pm$ 0.02 \\
                                 & bloom-7b1 & 32.20 $\pm$ 0.04                        & 16.00 $\pm$ 0.03                        & 0.472 $\pm$ 0.03                        \\
                                 & llama-7b  & 43.10 $\pm$ 0.03                      & 19.90 $\pm$ 0.04                         & 0.427 $\pm$ 0.03                        \\
                                 & llama-13b  &  36.30 $\pm$ 0.04                       & 17.70 $\pm$ 0.02                        & 0.460 $\pm$ 0.03                          \\
                                 & gpt-3.5   & \textbf{59.30} $\pm$ 0.04                        & \textbf{34.60} $\pm$ 0.02                         & \textbf{0.305} $\pm$ 0.03                        \\

\cmidrule(lr){1-5}
\multirow{7}{*}{5-item ranking}  & random guess  & 20.00 $\pm$ 0.00 & 0.83 $\pm$ 
                                  0.00  & 0.500 $\pm$ 0.00 \\
                                 & google stat  & 27.30 $\pm$ 0.01 & 3.30 $\pm$ 0.01  & 0.419 $\pm$ 0.02 \\
                                 & wiki len  & 46.70 $\pm$ 0.04 & 2.30 $\pm$ 0.01  & 0.337 $\pm$ 0.01 \\
                                 & bloom-7b1 & 17.90 $\pm$ 0.02                       & 0.50 $\pm$ 0.01                       & 0.473 $\pm$ 0.01                        \\
                                 & llama-7b  & 27.80 $\pm$ 0.02                       & 1.10 $\pm$ 0.01                         & 0.461 $\pm$ 0.01                       \\
                                 & llama-13b  & 24.90 $\pm$ 0.03                       & 1.60 $\pm$ 0.01                          & 0.446 $\pm$ 0.02                         \\
                                 & gpt-3.5   & \textbf{62.00} $\pm$ 0.03                        & \textbf{6.60} $\pm$ 0.00                       & \textbf{0.267} $\pm$ 0.02\tabularnewline

\bottomrule                       
\end{tabular}
}
\vspace{-0.5em}
\caption{Results of ranking worldwide holiday popularity in the US (mean $\pm$ standard deviation).}
\vspace{-0.5em}
\label{tab:llm_holiday}
\end{table}

\paragraph{Can LLMs elicit long-tail cultural statistics and rankings?} Table \ref{tab:llm_holiday} shows the ranking results of LLMs regarding holiday popularity in the US (see UK results in Appendix \ref{sec:ukresult}). In general, GPT-3.5 and LLaMa (7B and 13B) significantly outperform the random baseline, while BLOOM-7b1 tends to underperform on rankings. Notably, GPT-3.5 shows an obvious enhancement in most cases, except for pairwise holiday comparisons. Both statistical baselines outperformed the random guess with a high margin. Interestingly, the wiki baseline shows the highest accuracy in pairwise ranking and beats LLAMA variants in all ranking cases. Our observations demonstrate that GPT-3.5 and LLaMa exhibit the potential to capture the popularity tendencies of long-tail cultural concepts for ranking. A detailed holiday description on Wikipedia shows a positive signal related to the holiday popularity.

\begin{figure}[t]
   \centering
     \includegraphics[width=0.38\textwidth, height=3.8cm] 
     {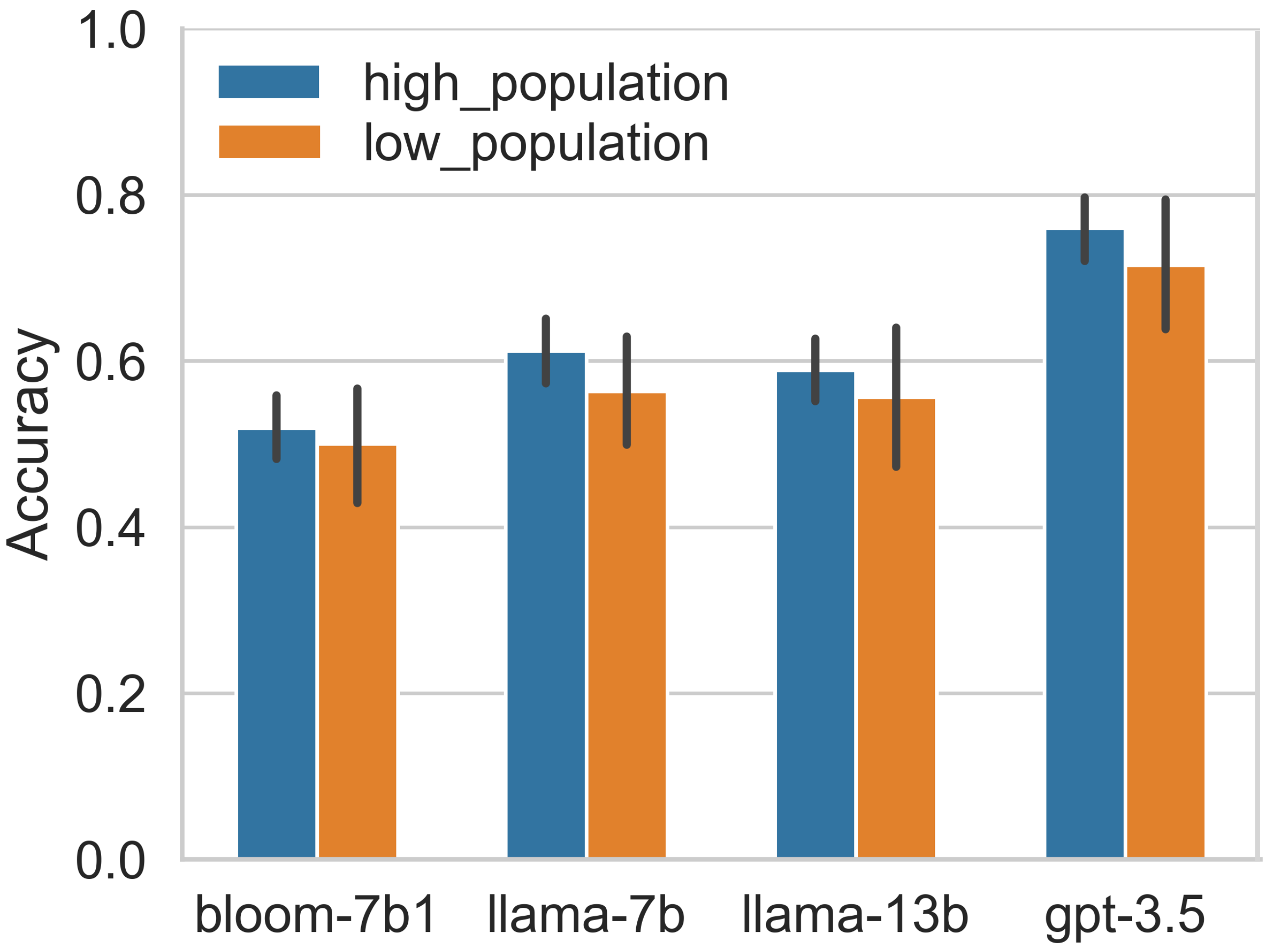}
     \vspace{-0.8ex}
     \caption{Pairwise ranking accuracy in the US regarding geo-cultural representativeness}
   \vspace{-1.5ex}
   \label{fig:us_country_pop}
\end{figure} 

\begin{figure}[t]
   \centering
     \includegraphics[width=0.35\textwidth, height=3.8cm] 
     {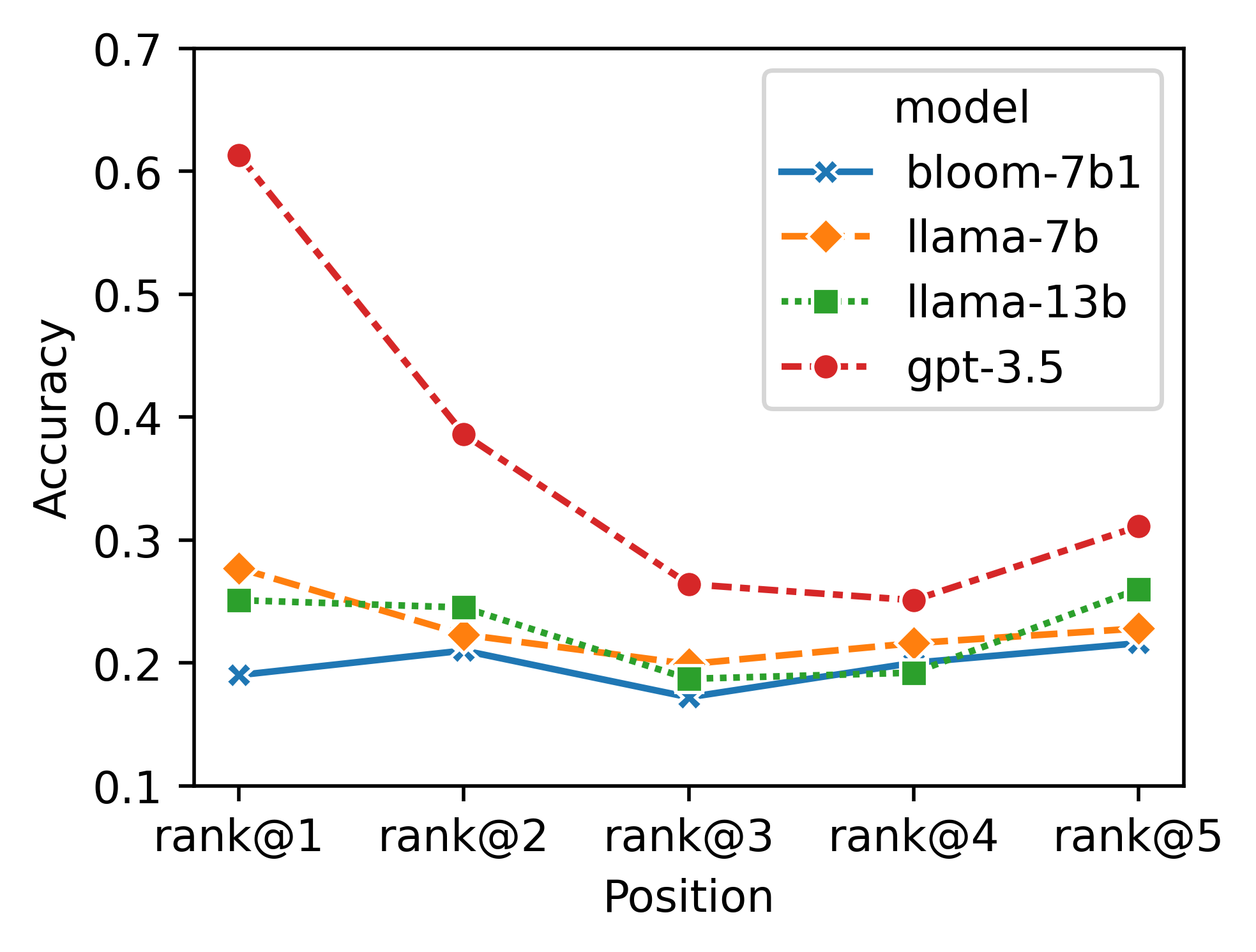}
     \vspace{-0.8ex}
     \caption{LLM results at each position on holiday ranking in the US}
   \vspace{-1.5ex}
   \label{fig:us_hpopularity}
\end{figure} 

\paragraph{What ranking-based factors challenge the prediction?} Looking into the ranking setting, we find that LLMs show a noticeable drop in P@1 ($\sim$5\%-10\%)and Acc ($\sim$20\%-30\%) when adding ranked items (see Table \ref{tab:llm_holiday}), suggesting that LLMs are sensitive to the ranking complexity. With further exploration of model performance at each ranking position (see Figure \ref{fig:us_hpopularity} on 5-item ranking results in the US, Appendix \ref{sec:ukresult} on results in the UK), we find that holidays ordered at two ends are usually easier to be predicted than those in between. Items ordered at the third and fourth positions are more prone to confuse LLMs than others.

\paragraph{Influence of geo-cultural representativeness?} To examine how LLMs respond to geo-cultural diversity, we conducted an analysis of pairwise ranking accuracy by comparing the most commonly shared holidays (by at least 10 countries) with country-specific holidays in high- versus low-population countries. As shown in Figure~\ref{fig:us_country_pop}, we observe that all models, particularly LLaMA-7b and GPT-3.5, exhibit higher accuracy when dealing with unique holidays from high-population countries as opposed to low-population ones. This outcome suggests that LLMs face challenges in capturing statistical trends related to under-represented geo-cultural concepts.

\paragraph{Possibility of LLMs embedding geo-cultural proximity?}
As countries with similar cultures tend to share common holidays, the cultural disparity between the query and target countries can influence LLM predictions. To investigate this, we group QA pairs by query locations and analyze model performance across distinct geo-groups. Given the dispersed distribution of QA pairs at the country level, we concentrate on continent-based comparisons using the optimal model, GPT-3.5. 

\begin{figure}[t]
   \centering
     \includegraphics[width=0.38\textwidth, height=4.2cm] 
     {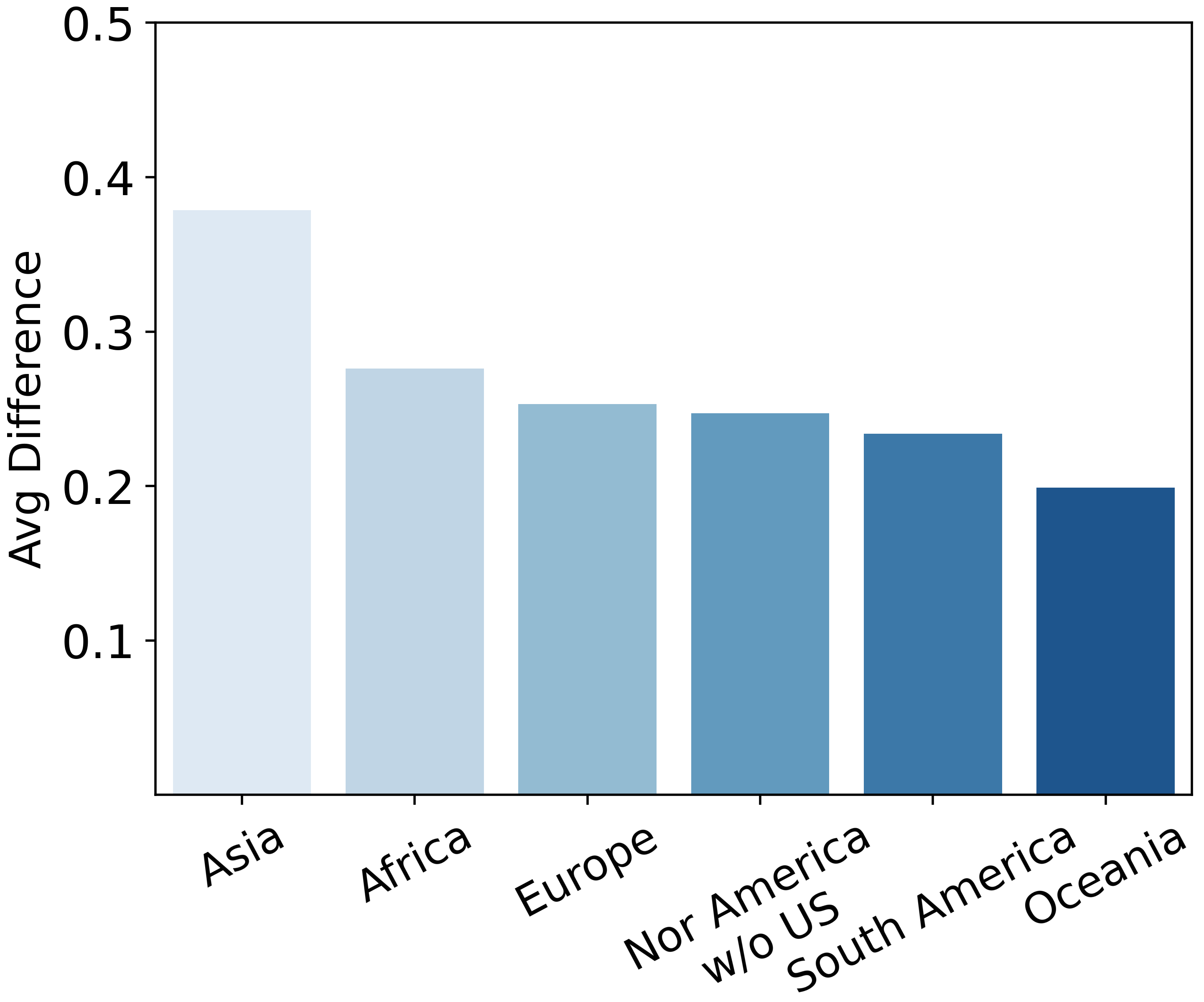}
     \vspace{-0.8ex}
     \caption{The results of GPT-3.5 on holiday ranking in the US across continents.}
     \vspace{-3.0ex}
   \label{fig: geo closeness}
\end{figure}

Figure \ref{fig: geo closeness} presents 5-item ranking results for holiday popularity in the US, while Appendix \ref{sec:ukresult} includes results for the UK. In Figure \ref{fig: geo closeness}, Oceania shows the lowest ranking differences, whereas Asia exhibits the highest. The predictability of GPT-3.5 aligns with geo-cultural proximity across continents, as seen in the Inglehart–Welzel cultural map \citet{culturalmap}. The Eastern culture dominant in Asia is more distant from the Western culture shared by Europe, Oceania, and the Americas. According to the cultural map, major Oceania countries (Australia and New Zealand) share a cultural group with the US. Despite the US being in North America, sampled non-US data in this continent mainly comes from low-population countries (e.g., Belize and Greenland), posing a potential challenge for GPT-3.5 predictions.

\section{Conclusion}
\label{sec:conclusion}

We introduce a novel QA task, CPopQA, to assess LLMs in ranking holiday-centered cultural concepts based on their popularity in the US and UK. Our results show that LLaMA and GPT-3.5 are capable of capturing implicit statistical tendencies of holiday popularity. Comparatively, GPT-3.5 displays superior ranking abilities. The model predictions are highly sensitive to ranking complexity in terms of the number of ranked items. Through further exploration of the optimal LLM (GPT-3.5), interestingly, we find that the model displays potential in identifying geo-cultural proximity across continents. By initiating the examination of LLMs' statistical ranking ability on long-tail cultural knowledge, this preliminary work benefits incentivizing future work on language models' cultural awareness.

\section{Limitations}
With a systematic review of our study, we summarize a list of limitations as follows.

First, regarding the holiday list, since we curated the holiday list based on Wikipedia, the potential data biases in Wikipedia such as missing holidays and countries, and misrepresentation of communities may cause issues of data representativeness in our dataset. Moreover, despite the diverse countries considered in this study, we focused on a sample of countries based on the accessible data from Wikipedia. The limited coverage of geo-political regions may also lead to unwanted data biases. 

Second, with respect to the holiday popularity collection, there may exist two concerns with the employment of Google Ngram Viewer to estimate holiday popularity. One is about the OCR quality of machine-digitized books, which may influence the n-gram statistical results. However, the tool developers have carefully considered this issue when building the tool \cite{ggngram1} and the later version further updated the OCR technology to improve the corpus quality \cite{ggngram2}. Considering the corpus in Google Ngram Viewer mainly consists of Google books, the other concern is about the domain shift issue. We will extend our study to consider diverse web resources for n-gram statistics in the future.

Third, in this preliminary study, we mainly focus on the use case of holiday popularity to investigate LLMs' potential on ranking-based statistical analysis questions. Moreover, the prompting template is simple as our study emphasizes the fundamental ability of LLMs in CPopQA. In the future, we will consider more diverse cultural concepts and a variety of prompting strategies for model evaluation.

\bibliography{main}

\begin{thebibliography}{43}
\expandafter\ifx\csname natexlab\endcsname\relax\def\natexlab#1{#1}\fi

\bibitem[{Acharya et~al.(2020)Acharya, Talamadupula, and Finlayson}]{culgeo2}
Anurag Acharya, Kartik Talamadupula, and Mark~A Finlayson. 2020.
\newblock Towards an atlas of cultural commonsense for machine reasoning.
\newblock \emph{arXiv preprint arXiv:2009.05664}.

\bibitem[{Arora et~al.(2023{\natexlab{a}})Arora, Kaffee, and
  Augenstein}]{cult2}
Arnav Arora, Lucie-aim{\'e}e Kaffee, and Isabelle Augenstein.
  2023{\natexlab{a}}.
\newblock Probing pre-trained language models for cross-cultural differences in
  values.
\newblock In \emph{Proceedings of the First Workshop on Cross-Cultural
  Considerations in NLP (C3NLP)}, pages 114--130, Dubrovnik, Croatia.
  Association for Computational Linguistics.

\bibitem[{Arora et~al.(2023{\natexlab{b}})Arora, Kaffee, and
  Augenstein}]{culprobing}
Arnav Arora, Lucie-aim{\'e}e Kaffee, and Isabelle Augenstein.
  2023{\natexlab{b}}.
\newblock Probing pre-trained language models for cross-cultural differences in
  values.
\newblock In \emph{Proceedings of the First Workshop on Cross-Cultural
  Considerations in NLP (C3NLP)}, pages 114--130, Dubrovnik, Croatia.

\bibitem[{Bosselut et~al.(2019)Bosselut, Rashkin, Sap, Malaviya, Celikyilmaz,
  and Choi}]{knowledgebase_1}
Antoine Bosselut, Hannah Rashkin, Maarten Sap, Chaitanya Malaviya, Asli
  Celikyilmaz, and Yejin Choi. 2019.
\newblock {COMET}: Commonsense transformers for automatic knowledge graph
  construction.
\newblock In \emph{Proceedings of the 57th Annual Meeting of the Association
  for Computational Linguistics}, pages 4762--4779, Florence, Italy.
  Association for Computational Linguistics.

\bibitem[{Breja and Jain(2022)}]{rankqa2}
Manvi Breja and Sanjay~Kumar Jain. 2022.
\newblock Analyzing linguistic features for answer re-ranking of why-questions.
\newblock \emph{Journal of Cases on Information Technology (JCIT)},
  24(3):1--16.

\bibitem[{Card et~al.(2022)Card, Chang, Becker, Mendelsohn, Voigt, Boustan,
  Abramitzky, and Jurafsky}]{dan_css}
Dallas Card, Serina Chang, Chris Becker, Julia Mendelsohn, Rob Voigt, Leah
  Boustan, Ran Abramitzky, and Dan Jurafsky. 2022.
\newblock Computational analysis of 140 years of us political speeches reveals
  more positive but increasingly polarized framing of immigration.
\newblock \emph{Proceedings of the National Academy of Sciences},
  119(31):e2120510119.

\bibitem[{Conneau et~al.(2018)Conneau, Kruszewski, Lample, Barrault, and
  Baroni}]{linguisticprob}
Alexis Conneau, German Kruszewski, Guillaume Lample, Lo{\"\i}c Barrault, and
  Marco Baroni. 2018.
\newblock What you can cram into a single {\$}{\&}!{\#}* vector: Probing
  sentence embeddings for linguistic properties.
\newblock In \emph{Proceedings of the 56th Annual Meeting of the Association
  for Computational Linguistics (Volume 1: Long Papers)}, pages 2126--2136,
  Melbourne, Australia. Association for Computational Linguistics.

\bibitem[{Garimella et~al.(2016)Garimella, Mihalcea, and Pennebaker}]{culword}
Aparna Garimella, Rada Mihalcea, and James Pennebaker. 2016.
\newblock Identifying cross-cultural differences in word usage.
\newblock In \emph{Proceedings of {COLING} 2016, the 26th International
  Conference on Computational Linguistics: Technical Papers}, pages 674--683,
  Osaka, Japan. The COLING 2016 Organizing Committee.

\bibitem[{Hershcovich et~al.(2022)Hershcovich, Frank, Lent, de~Lhoneux, Abdou,
  Brandl, Bugliarello, Cabello~Piqueras, Chalkidis, Cui, Fierro, Margatina,
  Rust, and S{\o}gaard}]{culnlp}
Daniel Hershcovich, Stella Frank, Heather Lent, Miryam de~Lhoneux, Mostafa
  Abdou, Stephanie Brandl, Emanuele Bugliarello, Laura Cabello~Piqueras, Ilias
  Chalkidis, Ruixiang Cui, Constanza Fierro, Katerina Margatina, Phillip Rust,
  and Anders S{\o}gaard. 2022.
\newblock Challenges and strategies in cross-cultural {NLP}.
\newblock In \emph{Proceedings of the 60th Annual Meeting of the Association
  for Computational Linguistics (Volume 1: Long Papers)}, pages 6997--7013,
  Dublin, Ireland. Association for Computational Linguistics.

\bibitem[{Hewitt and Manning(2019)}]{llmsyn}
John Hewitt and Christopher~D. Manning. 2019.
\newblock {A} structural probe for finding syntax in word representations.
\newblock In \emph{Proceedings of the 2019 Conference of the North {A}merican
  Chapter of the Association for Computational Linguistics: Human Language
  Technologies, Volume 1 (Long and Short Papers)}, pages 4129--4138,
  Minneapolis, Minnesota. Association for Computational Linguistics.

\bibitem[{Horawalavithana et~al.(2022)Horawalavithana, Ayton, Sharma, Howland,
  Subramanian, Vasquez, Cosbey, Glenski, and Volkova}]{science}
Sameera Horawalavithana, Ellyn Ayton, Shivam Sharma, Scott Howland, Megha
  Subramanian, Scott Vasquez, Robin Cosbey, Maria Glenski, and Svitlana
  Volkova. 2022.
\newblock Foundation models of scientific knowledge for chemistry:
  Opportunities, challenges and lessons learned.
\newblock In \emph{Proceedings of BigScience Episode {\#}5 -- Workshop on
  Challenges {\&} Perspectives in Creating Large Language Models}, pages
  160--172, virtual+Dublin. Association for Computational Linguistics.

\bibitem[{Hovy and Yang(2021)}]{socialfactor}
Dirk Hovy and Diyi Yang. 2021.
\newblock The importance of modeling social factors of language: Theory and
  practice.
\newblock In \emph{Proceedings of the 2021 Conference of the North American
  Chapter of the Association for Computational Linguistics: Human Language
  Technologies}, pages 588--602, Online. Association for Computational
  Linguistics.

\bibitem[{Imani et~al.(2023)Imani, Du, and Shrivastava}]{math}
Shima Imani, Liang Du, and Harsh Shrivastava. 2023.
\newblock Mathprompter: Mathematical reasoning using large language models.
\newblock \emph{arXiv preprint arXiv:2303.05398}.

\bibitem[{Inglehart and Welzel(2010)}]{culturalmap}
Ronald Inglehart and Chris Welzel. 2010.
\newblock The wvs cultural map of the world.
\newblock \emph{World Values Survey}.

\bibitem[{Jurgens et~al.(2017)Jurgens, Tsvetkov, and Jurafsky}]{culfeature}
David Jurgens, Yulia Tsvetkov, and Dan Jurafsky. 2017.
\newblock Incorporating dialectal variability for socially equitable language
  identification.
\newblock In \emph{Proceedings of the 55th Annual Meeting of the Association
  for Computational Linguistics (Volume 2: Short Papers)}, pages 51--57,
  Vancouver, Canada. Association for Computational Linguistics.

\bibitem[{Kabra et~al.(2023)Kabra, Liu, Khanuja, Aji, Winata, Cahyawijaya,
  Aremu, Ogayo, and Neubig}]{cultural}
Anubha Kabra, Emmy Liu, Simran Khanuja, Alham~Fikri Aji, Genta~Indra Winata,
  Samuel Cahyawijaya, Anuoluwapo Aremu, Perez Ogayo, and Graham Neubig. 2023.
\newblock Multi-lingual and multi-cultural figurative language understanding.
\newblock \emph{arXiv preprint arXiv:2305.16171}.

\bibitem[{Kandpal et~al.(2022)Kandpal, Deng, Roberts, Wallace, and
  Raffel}]{longtail}
Nikhil Kandpal, Haikang Deng, Adam Roberts, Eric Wallace, and Colin Raffel.
  2022.
\newblock Large language models struggle to learn long-tail knowledge.
\newblock \emph{arXiv preprint arXiv:2211.08411}.

\bibitem[{Kratzwald et~al.(2019)Kratzwald, Eigenmann, and
  Feuerriegel}]{rankqa3}
Bernhard Kratzwald, Anna Eigenmann, and Stefan Feuerriegel. 2019.
\newblock {R}ank{QA}: Neural question answering with answer re-ranking.
\newblock In \emph{Proceedings of the 57th Annual Meeting of the Association
  for Computational Linguistics}, pages 6076--6085, Florence, Italy.
  Association for Computational Linguistics.

\bibitem[{Li et~al.(2022)Li, Kuncoro, Hoffmann, de~Masson~d{'}Autume, Blunsom,
  and Nematzadeh}]{commonsense1}
Xiang~Lorraine Li, Adhiguna Kuncoro, Jordan Hoffmann, Cyprien
  de~Masson~d{'}Autume, Phil Blunsom, and Aida Nematzadeh. 2022.
\newblock A systematic investigation of commonsense knowledge in large language
  models.
\newblock In \emph{Proceedings of the 2022 Conference on Empirical Methods in
  Natural Language Processing}, pages 11838--11855, Abu Dhabi, United Arab
  Emirates. Association for Computational Linguistics.

\bibitem[{Lin et~al.(2012)Lin, Michel, Aiden~Lieberman, Orwant, Brockman, and
  Petrov}]{ggngram2}
Yuri Lin, Jean-Baptiste Michel, Erez Aiden~Lieberman, Jon Orwant, Will
  Brockman, and Slav Petrov. 2012.
\newblock Syntactic annotations for the {G}oogle {B}ooks {NG}ram corpus.
\newblock In \emph{Proceedings of the {ACL} 2012 System Demonstrations}, pages
  169--174, Jeju Island, Korea. Association for Computational Linguistics.

\bibitem[{Liu et~al.(2021)Liu, Bugliarello, Ponti, Reddy, Collier, and
  Elliott}]{culgeo1}
Fangyu Liu, Emanuele Bugliarello, Edoardo~Maria Ponti, Siva Reddy, Nigel
  Collier, and Desmond Elliott. 2021.
\newblock Visually grounded reasoning across languages and cultures.
\newblock In \emph{Proceedings of the 2021 Conference on Empirical Methods in
  Natural Language Processing}, pages 10467--10485, Online and Punta Cana,
  Dominican Republic. Association for Computational Linguistics.

\bibitem[{Meister and Cotterell(2021)}]{lmstatis1}
Clara Meister and Ryan Cotterell. 2021.
\newblock Language model evaluation beyond perplexity.
\newblock In \emph{Proceedings of the 59th Annual Meeting of the Association
  for Computational Linguistics and the 11th International Joint Conference on
  Natural Language Processing (Volume 1: Long Papers)}, pages 5328--5339,
  Online. Association for Computational Linguistics.

\bibitem[{Michel et~al.(2011)Michel, Shen, Aiden, Veres, Gray, Team, Pickett,
  Hoiberg, Clancy, Norvig et~al.}]{ggngram1}
Jean-Baptiste Michel, Yuan~Kui Shen, Aviva~Presser Aiden, Adrian Veres,
  Matthew~K Gray, Google~Books Team, Joseph~P Pickett, Dale Hoiberg, Dan
  Clancy, Peter Norvig, et~al. 2011.
\newblock Quantitative analysis of culture using millions of digitized books.
\newblock \emph{Science}, 331(6014):176--182.

\bibitem[{Nakov et~al.(2017)Nakov, Hoogeveen, M{\`a}rquez, Moschitti, Mubarak,
  Baldwin, and Verspoor}]{rankqa1}
Preslav Nakov, Doris Hoogeveen, Llu{\'\i}s M{\`a}rquez, Alessandro Moschitti,
  Hamdy Mubarak, Timothy Baldwin, and Karin Verspoor. 2017.
\newblock {S}em{E}val-2017 task 3: Community question answering.
\newblock In \emph{Proceedings of the 11th International Workshop on Semantic
  Evaluation ({S}em{E}val-2017)}, pages 27--48, Vancouver, Canada. Association
  for Computational Linguistics.

\bibitem[{Ouyang et~al.(2022)Ouyang, Wu, Jiang, Almeida, Wainwright, Mishkin,
  Zhang, Agarwal, Slama, Ray et~al.}]{gpt35}
Long Ouyang, Jeffrey Wu, Xu~Jiang, Diogo Almeida, Carroll Wainwright, Pamela
  Mishkin, Chong Zhang, Sandhini Agarwal, Katarina Slama, Alex Ray, et~al.
  2022.
\newblock Training language models to follow instructions with human feedback.
\newblock \emph{Advances in Neural Information Processing Systems},
  35:27730--27744.

\bibitem[{Petroni et~al.(2019)Petroni, Rockt{\"a}schel, Riedel, Lewis, Bakhtin,
  Wu, and Miller}]{factual1}
Fabio Petroni, Tim Rockt{\"a}schel, Sebastian Riedel, Patrick Lewis, Anton
  Bakhtin, Yuxiang Wu, and Alexander Miller. 2019.
\newblock Language models as knowledge bases?
\newblock In \emph{Proceedings of the 2019 Conference on Empirical Methods in
  Natural Language Processing and the 9th International Joint Conference on
  Natural Language Processing (EMNLP-IJCNLP)}, pages 2463--2473, Hong Kong,
  China. Association for Computational Linguistics.

\bibitem[{Riley et~al.(2022)Riley, Dozat, Botha, Garcia, Garrette, Riesa,
  Firat, and Constant}]{culmt}
Parker Riley, Timothy Dozat, Jan~A Botha, Xavier Garcia, Dan Garrette, Jason
  Riesa, Orhan Firat, and Noah Constant. 2022.
\newblock Frmt: A benchmark for few-shot region-aware machine translation.
\newblock \emph{arXiv preprint arXiv:2210.00193}.

\bibitem[{Ringel et~al.(2019)Ringel, Lavee, Guy, and Radinsky}]{culclass}
Dor Ringel, Gal Lavee, Ido Guy, and Kira Radinsky. 2019.
\newblock Cross-cultural transfer learning for text classification.
\newblock In \emph{Proceedings of the 2019 Conference on Empirical Methods in
  Natural Language Processing and the 9th International Joint Conference on
  Natural Language Processing (EMNLP-IJCNLP)}, pages 3873--3883, Hong Kong,
  China. Association for Computational Linguistics.

\bibitem[{Roberts(2020)}]{statistical_text}
Carl~W Roberts. 2020.
\newblock \emph{Text analysis for the social sciences: methods for drawing
  statistical inferences from texts and transcripts}.
\newblock Routledge.

\bibitem[{Rogers et~al.(2023)Rogers, Gardner, and Augenstein}]{qadatasurvey}
Anna Rogers, Matt Gardner, and Isabelle Augenstein. 2023.
\newblock Qa dataset explosion: A taxonomy of nlp resources for question
  answering and reading comprehension.
\newblock \emph{ACM Computing Surveys}, 55(10):1--45.

\bibitem[{Scao et~al.(2023)Scao, Fan, Akiki, Pavlick, Ilić, Hesslow,
  Castagné, Luccioni, Yvon, Gallé, Tow, Rush, and et~al.}]{bloom}
Teven~Le Scao, Angela Fan, Christopher Akiki, Ellie Pavlick, Suzana Ilić,
  Daniel Hesslow, Roman Castagné, Alexandra~Sasha Luccioni, François Yvon,
  Matthias Gallé, Jonathan Tow, Alexander~M. Rush, and et~al. 2023.
\newblock Bloom: A 176b-parameter open-access multilingual language model.
\newblock \emph{arXiv preprint arXiv:2211.05100}.

\bibitem[{Shinn et~al.(2023)Shinn, Cassano, Labash, Gopinath, Narasimhan, and
  Yao}]{selfreflexion}
Noah Shinn, Federico Cassano, Beck Labash, Ashwin Gopinath, Karthik Narasimhan,
  and Shunyu Yao. 2023.
\newblock Reflexion: Language agents with verbal reinforcement learning.

\bibitem[{Soni et~al.(2023)Soni, Sihra, Evans, Wilkens, and
  Bamman}]{humanities}
Sandeep Soni, Amanpreet Sihra, Elizabeth~F. Evans, Matthew Wilkens, and David
  Bamman. 2023.
\newblock Grounding characters and places in narrative texts.
\newblock \emph{arXiv preprint arXiv:2305.17561}.

\bibitem[{Sun et~al.(2021)Sun, Ahn, Park, Tsvetkov, and
  Mortensen}]{culturalsim}
Jimin Sun, Hwijeen Ahn, Chan~Young Park, Yulia Tsvetkov, and David~R.
  Mortensen. 2021.
\newblock Cross-cultural similarity features for cross-lingual transfer
  learning of pragmatically motivated tasks.
\newblock In \emph{Proceedings of the 16th Conference of the European Chapter
  of the Association for Computational Linguistics: Main Volume}, pages
  2403--2414, Online. Association for Computational Linguistics.

\bibitem[{Takahashi and Tanaka-Ishii(2017)}]{llmstatis2}
Shuntaro Takahashi and Kumiko Tanaka-Ishii. 2017.
\newblock Do neural nets learn statistical laws behind natural language?
\newblock \emph{PloS one}, 12(12):e0189326.

\bibitem[{Takahashi and Tanaka-Ishii(2019)}]{llmstatis3}
Shuntaro Takahashi and Kumiko Tanaka-Ishii. 2019.
\newblock Evaluating computational language models with scaling properties of
  natural language.
\newblock \emph{Computational Linguistics}, 45(3):481--513.

\bibitem[{Touvron et~al.(2023)Touvron, Lavril, Izacard, Martinet, Lachaux,
  Lacroix, Rozi{\`e}re, Goyal, Hambro, Azhar et~al.}]{llama}
Hugo Touvron, Thibaut Lavril, Gautier Izacard, Xavier Martinet, Marie-Anne
  Lachaux, Timoth{\'e}e Lacroix, Baptiste Rozi{\`e}re, Naman Goyal, Eric
  Hambro, Faisal Azhar, et~al. 2023.
\newblock Llama: Open and efficient foundation language models.
\newblock \emph{arXiv preprint arXiv:2302.13971}.

\bibitem[{Underwood and So(2021)}]{culture_ted}
Ted Underwood and Richard~Jean So. 2021.
\newblock Can we map culture?
\newblock \emph{Journal of Cultural Analytics}, 6(3).

\bibitem[{Wang et~al.(2022)Wang, Deng, and Sun}]{cot3}
Boshi Wang, Xiang Deng, and Huan Sun. 2022.
\newblock Iteratively prompt pre-trained language models for chain of thought.
\newblock In \emph{Proceedings of the 2022 Conference on Empirical Methods in
  Natural Language Processing}, pages 2714--2730.

\bibitem[{Wei et~al.(2022)Wei, Wang, Schuurmans, Bosma, Chi, Le, and
  Zhou}]{chainofthought}
Jason Wei, Xuezhi Wang, Dale Schuurmans, Maarten Bosma, Ed~Chi, Quoc Le, and
  Denny Zhou. 2022.
\newblock Chain of thought prompting elicits reasoning in large language
  models.
\newblock \emph{arXiv preprint arXiv:2201.11903}.

\bibitem[{Wei et~al.(2023)Wei, Cui, Cheng, Wang, Zhang, Huang, Xie, Xu, Chen,
  Zhang et~al.}]{knowledgebase_2}
Xiang Wei, Xingyu Cui, Ning Cheng, Xiaobin Wang, Xin Zhang, Shen Huang, Pengjun
  Xie, Jinan Xu, Yufeng Chen, Meishan Zhang, et~al. 2023.
\newblock Zero-shot information extraction via chatting with chatgpt.
\newblock \emph{arXiv preprint arXiv:2302.10205}.

\bibitem[{Yin et~al.(2022)Yin, Bansal, Monajatipoor, Li, and
  Chang}]{geocommonsense}
Da~Yin, Hritik Bansal, Masoud Monajatipoor, Liunian~Harold Li, and Kai-Wei
  Chang. 2022.
\newblock {G}eo{MLAMA}: Geo-diverse commonsense probing on multilingual
  pre-trained language models.
\newblock In \emph{Proceedings of the 2022 Conference on Empirical Methods in
  Natural Language Processing}, pages 2039--2055, Abu Dhabi, United Arab
  Emirates. Association for Computational Linguistics.

\bibitem[{Zhang et~al.(2021)Zhang, Zhang, Zhang, and S{\o}gaard}]{culdialect}
Sheng Zhang, Xin Zhang, Weiming Zhang, and Anders S{\o}gaard. 2021.
\newblock Sociolectal analysis of pretrained language models.
\newblock In \emph{Proceedings of the 2021 Conference on Empirical Methods in
  Natural Language Processing}, pages 4581--4588, Online and Punta Cana,
  Dominican Republic. Association for Computational Linguistics.

\end{thebibliography}

\newpage
\appendix
\onecolumn

\label{sec:appendix}

\section{Related Work}
\paragraph{Cultural-aware NLP} Language and culture are intertwined \cite{culnlp,socialfactor}. Overall, research on the interaction of language technologies and cultures can be divided into two groups. The first group focuses on improving specific language technologies inspired by cultural diversity \cite{culturalsim,culfeature,culmt}. For example, \citet{culfeature} proposed a new distance measure between languages based on linguistic proxies of culture, hoping to improve cross-lingual transfer learning. \citet{culmt} constructed a benchmark called FRMT to improve matching translation with an emphasis on geo-cultural diversity. 

The second group concentrates on investigating the cultural awareness of language technologies \cite{culprobing,culclass,culword}. Popular research topics in this thread include cross-cultural differences in word usage \cite{culword}, dialect-associated biases \cite{culdialect}, and geo-diverse commonsense \cite{culgeo1,culgeo2,geocommonsense}.

\paragraph{Ranking-based QA} Existing work in the field of ranking-based QA primarily focus on answer re-ranking to identify the optimal one \cite{rankqa1,rankqa2,rankqa3}. Following \citet{qadatasurvey}, one of the major motivations behind this group of studies lies in the diversity in both the quality and quantity of questions and answers. Differing from prior studies that focus on developing ranking-based QA models to identify the best answer from a pool of candidates, our study specifically centers around a QA task that aims to generate a ranking of cultural concepts (holidays) based on their popularity.

\section{Data Cleaning}
\label{sec:datacleaning}
We conducted both rule-based cleaning and post-human edition  to improve the data quality. Specifically, we filtered out holidays that lose the time description for further consideration. Regarding temporal diversity, we employed two human annotators to unify the holiday date following Gregorian Calendar. Considering the temporal dynamics of some holidays caused by calendar conversion, we further required annotators to assign the label "movable" to these holidays' dates. Moreover, the paraphrase phenomenon of some holidays may cause their popularity distribution to be dispersed. To avoid this issue, we examined each holiday concept and grouped its aliases. In addition, through the empirical examination of the holiday list, annotators also removed false positives (e.g., special events like the memory of an emperor) and improved holiday descriptions by manual edition.

\section{Human Evaluation of Holiday Popularity Collection}
\label{sec:humaneval}
To further validate our strategy for deriving holiday popularity, we additionally conducted a human evaluation of holiday popularity rankings. Specifically, we randomly sampled 5 countries and selected the top 10 holidays per country based on their frequency in GBNV’s American English corpus. For each country’s holiday list, we asked 6 non-immigrant US citizens, who grew up in the US, to compare holidays regarding their popularity in the US and generated a rank based on annotators’ average votes. Toward a correlation analysis of two ranked holiday lists per country, our results show that the statistics of Google Books Ngram Viewer achieved ~60\% consistency (i.e., Pearson p=63.34\%, Spearman rho=58.65\%) with human judgments on average.

\clearpage
\begin{figure*}[h!]
   \centering
     \includegraphics[width=0.98\textwidth] 
     {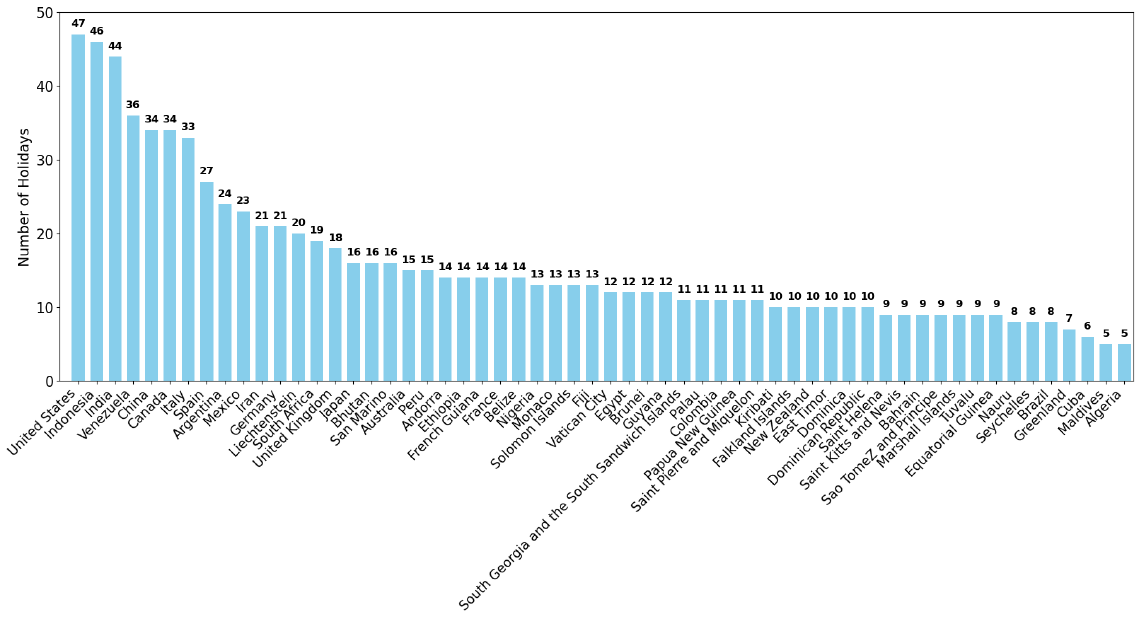}
     \caption{Distribution of holidays by country in descending order.}
   \label{fig:holiday_country}
\end{figure*} 

\begin{figure}[!htb]
   \centering
     \includegraphics[width=0.50\textwidth] 
     {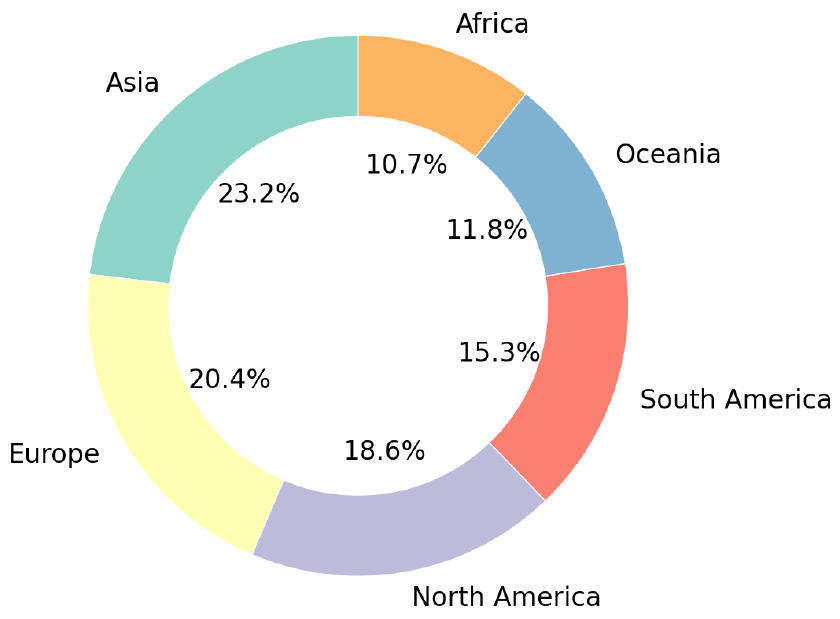}
     \caption{Distribution of holidays by continent.}
   \label{fig:holiday_continents_doghnutChart}
\end{figure} 

\section{Holiday Statistics}
\label{sec:datastat}
Figure \ref{fig:holiday_country} displays the distribution of holidays across various countries. Our dataset comprises a total of 459 unique holidays in 58 countries. Notably, the United States, Indonesia, and India are the top three countries with the highest number of holidays. Conversely, Cuba, Maldives, and Algeria have the lowest number of holidays among the countries included in our dataset. 

Figure \ref{fig:holiday_continents_doghnutChart} presents the distribution of holidays by continent. In comparison, Asia (23.2\%) and Europe  (20.4\%) emerge as the top two continents with a higher number of holidays compared to others. On the other hand, Africa (10.7\%) and Oceania (11.8\%) have a relatively lower ratio of holidays in comparison to the other continents.

\clearpage
\section{Details of Baselines}
\label{sec:baseline}
Regarding baselines, in addition to widely used random guess, we proposed two additional baselines. For the first baseline which we called “google stat”, we quantified the general holiday popularity by querying each holiday in a search engine and estimating its cumulative search volume over time. Given that Google is one of the most popular search engines, we used Google Trends (\url{https://trends.google.com/trends/}) to access the search volume of each holiday query in Google Search across ~20 years (2004-1-1 - 2023-8-1, the maximum accessible timeline in the tool) and sum up the statistics over the selected time span. The second baseline is called “wiki len”. With the assumption that a well-known holiday tends to contain a more comprehensive and lengthy description compared to a lesser-known one, we calculated the word length of the description on each holiday’s Wikipedia page to approximate the holiday’s popularity.

\section{Detailed Examples of QA Pairs}
In Table~\ref{tab:us_exps}, we provide a few detailed examples of QA pairs for holiday ranking. Note that, we don't use the popularity information or holiday descriptions in our prompts. However, such information can be served as valuable context for future studies. 

\subsection{Holiday Ranking}
\begin{table}[!h]
\centering
\resizebox{0.98\columnwidth}{!}{%
\small
\begin{tabular}{p{0.3cm}|p{1.5cm}p{11cm}}
\toprule
No.                 & Attribute          & Content   \\                                            \midrule                                                                                                                                                                                   
                    & Question           & Can you provide a descending order for the following \textcolor{blue}{Chinese} holidays by their popularity in the \textcolor{orange}{United States}: Spring Festival, Children's Day, Lantern Festival, Martyrs' Day, Dragon Boat Festival?                                     \\
                    \cmidrule(lr){2-3}
                    & Answer             & 1. Spring Festival 2. Children's Day 3. Dragon Boat Festival 4. Lantern Festival 5. Martyrs' Day 
                    \\
                    \cmidrule(lr){2-3}
                    & Holiday Popularity & The percentage of the frequency of the holiday Chinese New Year / Spring Festival mentioned in the United States corpus is 1.11e-07. The percentage of ...                                                                                 \\
                    \cmidrule(lr){2-3}
\multirow{-4}{*}{1} & Description        & Chinese New Year / Spring Festival: Chinese New Year is the festival that celebrates the beginning of a new year on the traditional lunisolar Chinese calendar. In Chinese, the festival is commonly referred to as the Spring Festival... \\
\midrule 
                    & Question           & Can you provide a descending order for the following \textcolor{blue}{Indian} holidays by their popularity in the \textcolor{orange}{United States}: Independence Day, Gandhi Jayanti, Bihu, Accession Day, Day of Ashura?                               \\
                    \cmidrule(lr){2-3}
                    & Answer             & 1. Independence Day 2. Accession Day 3. Bihu 4. Day of Ashura 5. Gandhi Jayanti                                                                                                                                                            \\
                    \cmidrule(lr){2-3}
                    & Holiday Popularity & The percentage of the frequency of the holiday Independence Day mentioned in the United States corpus is 4.07e-07. The percentage of ...                                                                                                   \\
                    \cmidrule(lr){2-3}
\multirow{-4}{*}{2} & Description        & Independence Day: Independence Day is celebrated annually on 15 August as a public holiday in India commemorating the nation's independence from the United Kingdom...                                                                     \\
\midrule 
                    & Question           & Can you provide a descending order for the following \textcolor{blue}{Fijian} holidays by their popularity in the \textcolor{orange}{United States}: New Year’s Day, Ram Naumi, Fiji Day, National Youth Day, Palm Sunday?                                                       \\
                    \cmidrule(lr){2-3}
                    & Answer             & 1. New Year's Day 2. Palm Sunday 3. National Youth Day 4. Fiji Day 5. Ram Naumi                                                                                                                                                            \\
                    \cmidrule(lr){2-3}
                    & Holiday Popularity & The percentage of the frequency of the holiday New Year's Day mentioned in the United States corpus is 5.78e-07. The percentage of ...                                                                                                     \\
                    \cmidrule(lr){2-3}
\multirow{-4}{*}{3} & Description        & New Year's Day: In the Gregorian calendar, New Year's Day is the first day of the year ...\tabularnewline 
\bottomrule   

\end{tabular}
}
\caption{QA pair examples for 5-holiday ranking in the US.}
\label{tab:us_exps}
\end{table}

\section{Consistency of Holiday Rank in US versus UK}
Table~\ref{tab:us-uk diff} displays the level of ranking consistency between the popularity of holidays in American culture compared to British culture. In particular, the agreement metric measures the extent of the exact match between the rank in the US and the rank in the UK. We also compute the average ranking difference across queries.

\begin{table}[!h]
    \centering
    \resizebox{0.45\columnwidth}{!}{%
    \begin{tabular}{c|cc}
    \toprule
     \multicolumn{1}{l|}{}  & \multicolumn{1}{l}{Agreement (\%)} & \multicolumn{1}{l}{Diff.} \\
    \midrule
    2-item ranking & 93.00 & -     \\
    3-item ranking & 75.80 & 0.073   \\
    5-item ranking & 39.20 & 0.061  \tabularnewline 
    \bottomrule   
    \end{tabular}
    }
    \caption{Consistency of holiday popularity rank in US versus UK.}
    \label{tab:us-uk diff}
\end{table}

\section{Results of LLMs on Holiday Ranking in the UK}
\label{sec:ukresult}
\begin{table}[!h]
\centering
\resizebox{0.65\columnwidth}{!}{%
\begin{tabular}{llccc}
\toprule
Setting                          & Model  & \multicolumn{1}{l}{P@1 (\%)} & \multicolumn{1}{l}{Acc. (\%)} & \multicolumn{1}{l}{Diff.}    \\
\midrule
\multirow{7}{*}{2-item ranking} & random guess  & 50.00 $\pm$ 0.00 & 50.00 $\pm$ 0.00 & -     \\
                                 & google stat  & 56.40 $\pm$ 0.03 & 56.40 $\pm$ 0.03  & -     \\
                                 & wiki len  & \textbf{61.30} $\pm$ 0.04 & \textbf{61.30} $\pm$ 0.04  & -     \\
                                 & bloom-7b1 & 39.40 $\pm$ 0.04                        & 39.40 $\pm$ 0.04                      & -                            \\
                                 & llama-7b  & 46.80 $\pm$ 0.02                     & 46.80 $\pm$ 0.02                         & -                            \\
                                 & llama-13b  & 49.50 $\pm$ 0.03                        & 49.50 $\pm$ 0.03                           & -                            \\
                                 & gpt-3.5   & 53.50 $\pm$ 0.04                       & 53.50 $\pm$ 0.04                        & -                            \\
\cmidrule(lr){1-5}
\multirow{7}{*}{3-item ranking} & random guess  & 33.33 $\pm$ 0.00 & 16.67 
                                   $\pm$ 0.00  & 0.500 $\pm$ 0.00 \\
                                 & google stat  & 36.30 $\pm$ 0.03 & 20.00 $\pm$ 0.02  & 0.448 $\pm$ 0.01 \\
                                 & wiki len  & 54.60 $\pm$ 0.03 & 29.80 $\pm$ 0.03  & 0.361 $\pm$ 0.02 \\
                                 & bloom-7b1 & 31.90 $\pm$ 0.01                        & 16.60 $\pm$ 0.03                        & 0.481 $\pm$ 0.01                        \\
                                 & llama-7b  & 37.60 $\pm$ 0.03                      & 17.00 $\pm$ 0.03                         & 0.466 $\pm$ 0.01                        \\
                                 & llama-13b  &  37.10 $\pm$ 0.03                       & 19.80 $\pm$ 0.04                        & 0.455 $\pm$ 0.02                          \\
                                 & gpt-3.5   & \textbf{62.60} $\pm$ 0.01                        & \textbf{38.80} $\pm$ 0.01                         & \textbf{0.278} $\pm$ 0.01                        \\

\cmidrule(lr){1-5}
\multirow{7}{*}{5-item ranking}  & random guess  & 20.00 $\pm$ 0.00 & 0.83 $\pm$ 
                                  0.00  & 0.500 $\pm$ 0.00 \\
                                 & google stat  & 23.80 $\pm$ 0.01 & 2.40 $\pm$ 0.01  & 0.431 $\pm$ 0.02 \\
                                 & wiki len  & 53.10 $\pm$ 0.04 & 2.90 $\pm$ 0.01  & 0.322 $\pm$ 0.01 \\
                                 & bloom-7b1 & 18.10 $\pm$ 0.03                       & 0.80 $\pm$ 0.01                       & 0.477 $\pm$ 0.02                        \\
                                 & llama-7b  & 29.50 $\pm$ 0.03                       & 1.50 $\pm$ 0.01                         & 0.462 $\pm$ 0.01                       \\
                                 & llama-13b  & 28.30 $\pm$ 0.03                       & 2.20 $\pm$ 0.00                          & 0.420 $\pm$ 0.03                         \\
                                 & gpt-3.5   & \textbf{60.60} $\pm$ 0.04                        & \textbf{7.60} $\pm$ 0.02                       & \textbf{0.258} $\pm$ 0.01\tabularnewline

\bottomrule    
\end{tabular}
}
\caption{Performance of LLMs on worldwide holiday popularity rankings in the UK (mean $\pm$ standard deviation).}
\label{tab:llm_uk_holiday_5rounds}
\end{table}

\begin{figure}[!h]
   \centering
     \includegraphics[width=0.65\textwidth] 
     {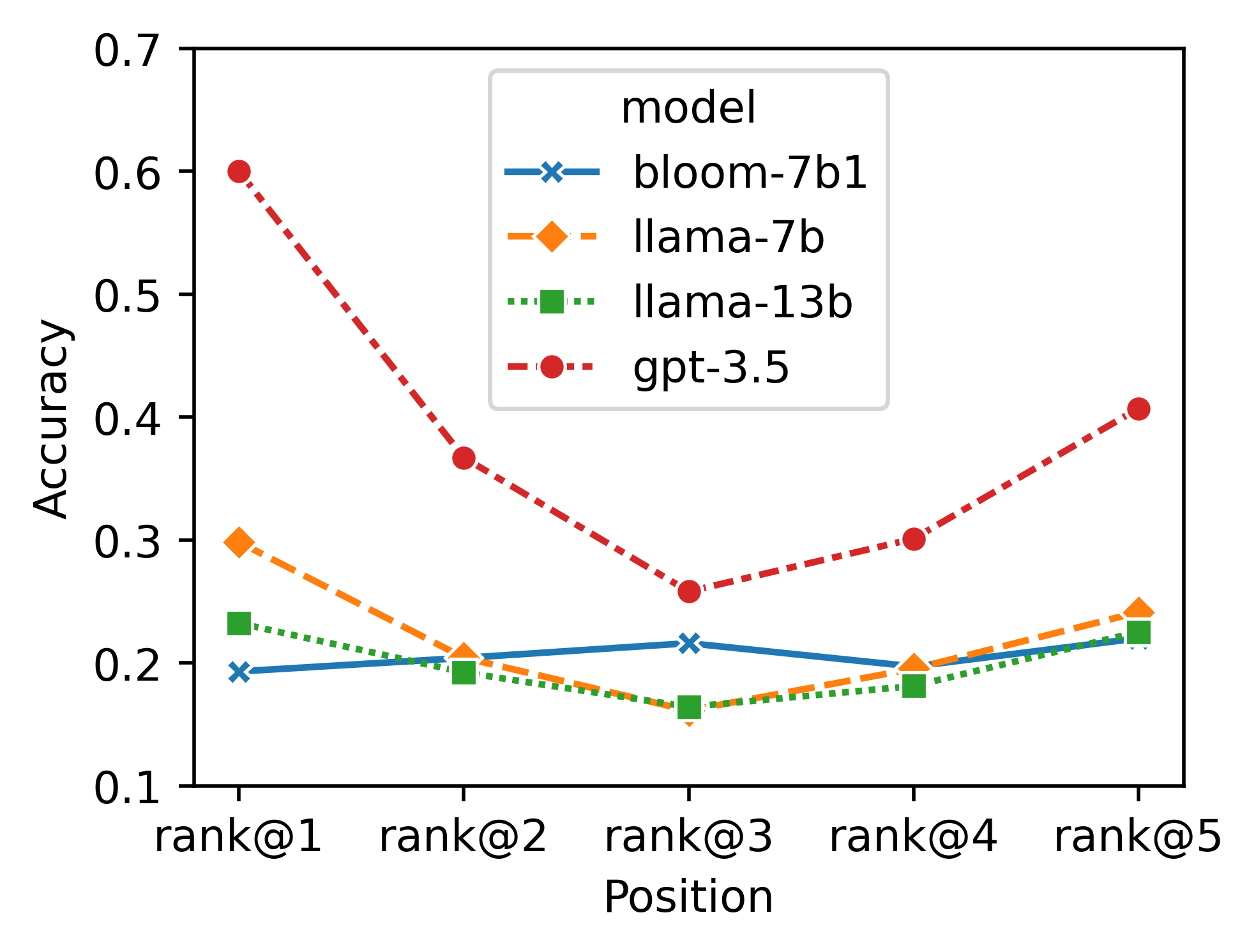}
     \caption{LLM results at each position on holiday ranking in the UK.}
   \label{fig:uk_hpopularity}
\end{figure}  

\begin{figure}[!hbt]
   \centering
     \includegraphics[width=0.65\textwidth] 
     {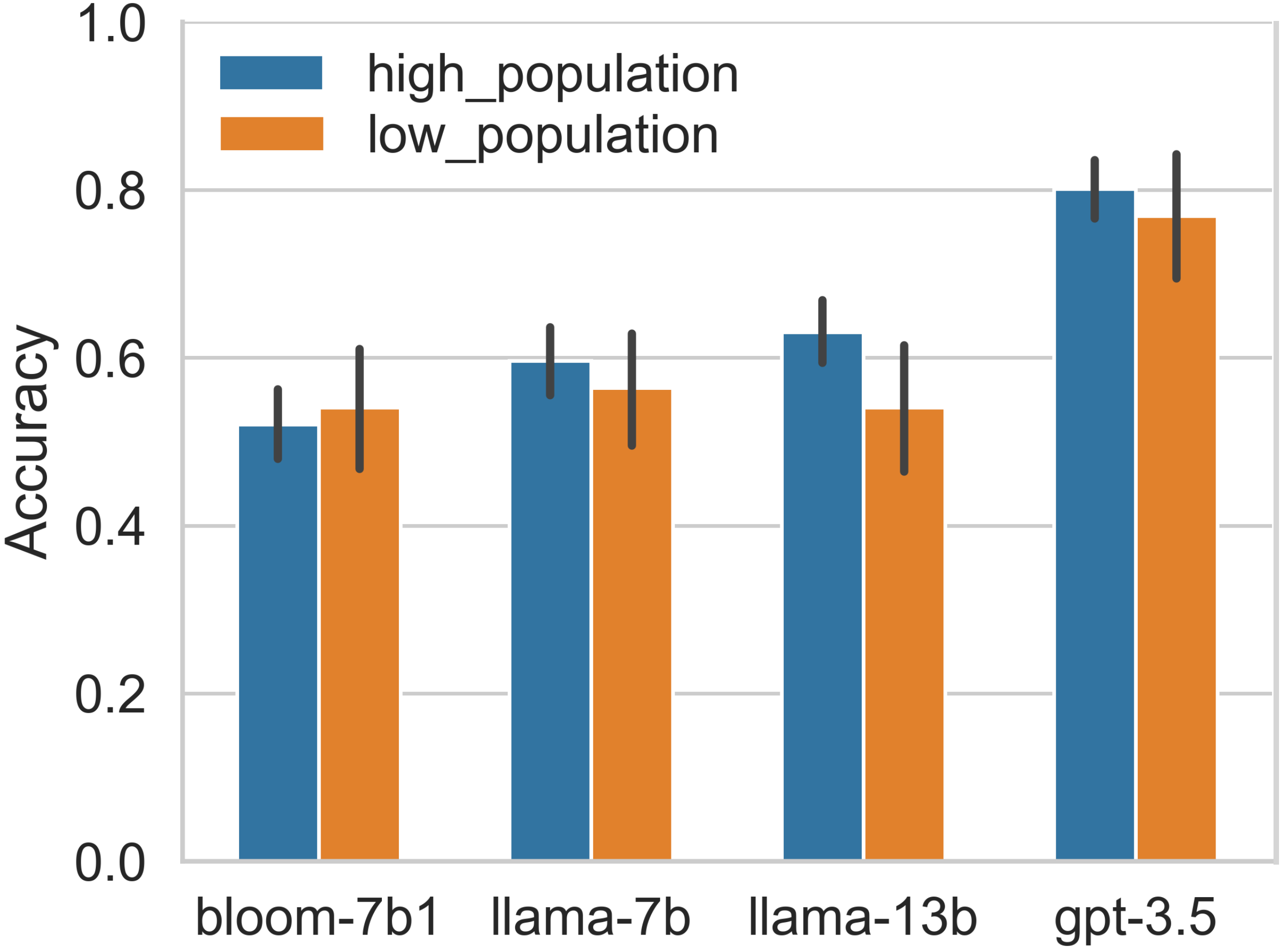}
     \vspace{-0.8ex}
     \caption{Pairwise ranking accuracy in the UK regarding geo-cultural representativeness}
   \vspace{-1.5ex}
   \label{fig:uk_country_pop}
\end{figure} 

\begin{figure}[!hbt]
   \centering
     \includegraphics[width=0.65\textwidth] 
     {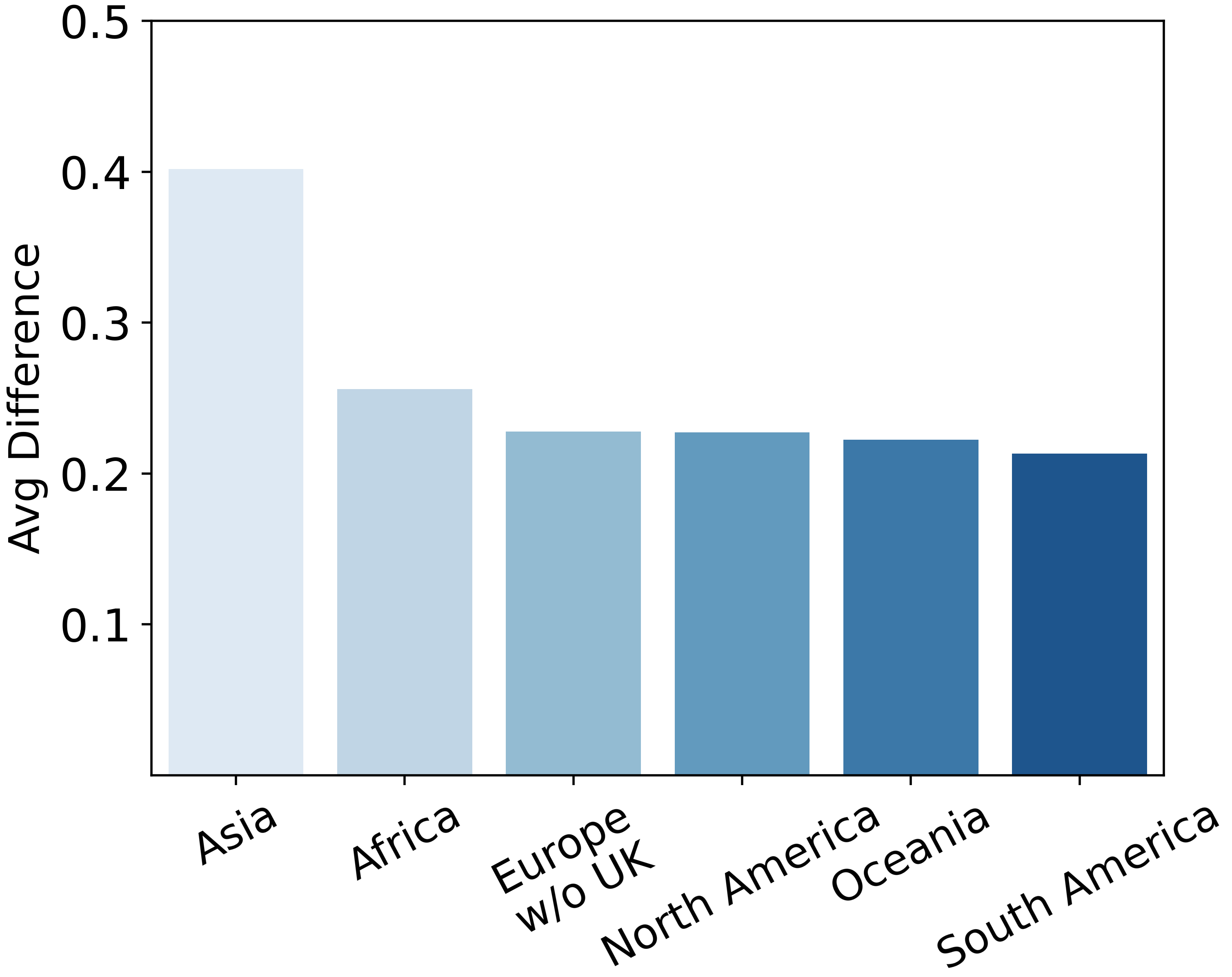}
     \caption{The results of GPT-3.5 on holiday ranking in the UK across continents.}
   \label{fig:uk_geo closeness}
\end{figure}

\end{document}